\documentclass{article}

\usepackage{arxiv}

\usepackage[utf8]{inputenc} 
\usepackage[T1]{fontenc}    
\usepackage{hyperref}       
\usepackage{url}            
\usepackage{booktabs}       
\usepackage{amsfonts}       
\usepackage{nicefrac}       
\usepackage{microtype}      
\usepackage{lipsum}

\usepackage{graphicx}
\usepackage{subcaption}

\title{Fake news detection using Deep Learning}

\author{
  Álvaro Ibrain Rodríguez \\
  Department of Computer Science\\
  University of Cantabria\\
  Spain \\
  \texttt{alvaroibrain@outlook.es } \\
   \And
 Lara Lloret Iglesias \\
  Department of Advanced Computing and e-science\\
  IFCA-CSIC-UC\\
  Spain \\
  \texttt{lloret@ifca.unican.es} \\
}

\begin{document}
\maketitle

\begin{abstract}
    The evolution of the information and communication technologies has dramatically 
    increased the number of people with access to the Internet, which has changed the
    way the information is consumed.

    As a consequence of the above, fake news have become one of the major concerns because
    its potential to destabilize governments, which makes them a potential danger to
    modern society. An example of this can be found in the US. electoral campaign,
    where the term ``\textit{fake news}'' gained great notoriety due to the influence
    of the hoaxes in the final result of these.

    In this work the feasibility of applying deep learning techniques to discriminate fake
    news on the Internet using only their text is studied. In order to accomplish that, three
    different neural network architectures are proposed, one of them based on BERT, a
    modern language model created by Google which achieves state-of-the-art results.
\end{abstract}


\section{Introduction}

The rise of new information technologies has produced many new ways of consumption like online shopping, multimedia entertainment, gaming, advertising or learning, among others.

One of the sectors greatly impacted by this new paradigm is the information industry. In the last years, there has been an exodus of users from the more traditional media such as newspapers, radio and television to new formats: social networks, YouTube, podcasts, online journals, news applications, etc~\cite{news-comsumption-trend}. The main cause of the information media decay 
is the growing capability, thanks to the Internet, of having instant and free access to a wide variety of information sources let alone numerous services that allow sharing the news with millions of people around the world.

As a result, the media have started to react to the change. Some, for example, have begun to prioritize their online presence or have decided to start using new distribution channels such as videos or podcasts. Their current project is finding a way of making profitable these new distribution formats that have always been free for the end user.

Most of these media have decided to start monetizing their content through advertising embedded in their articles, videos, etc. One of the most frequent methods is publishing articles with flashy headlines and photos aimed to be shared on social networks (known as \textit{clickbait}) so that users
navigate to their websites thus maximizing their revenue.

However, this kind of approach can lead to dangerous situations. The large amount of information that people access
is usually unverified and generally assumed as true \cite{socialmedia-and-fake-news-2016}.

It is at this point where the term fake news arises. This problem has reached its peak of notoriety after the 2016 US electoral campaign \cite{trump-fakenews} when it has been determined that the fake news have been crucial to polarize society and promote the triumph of Donald Trump \cite{trump-not-won-fknews}.

The big technology companies (Twitter, Facebook, Google) have spotted
this danger and have already begun to work on systems to detect these fake news on their platforms. All the same, even if this methods evolve fast, this is still a very complicated problem that needs further investigation.

The main objective of this work is to obtain several models based on deep neural networks aimed at detecting and categorizing the fake news so that 
people can have some trust degree indicator of the information they consume and thus avoid, to the extent possible, bias and misconceptions.

\subsection{Problem definition}

To express the problem to solve in a more formal way; given $a$ as a news article defined by a set of own characteristics (ie. title, text, photos, newspaper, author, ...), a function $f$ is sought such as:
\begin{equation}
	f(a) = \cases{
				0 & if $a$ is fake \cr
				1 & if $a$ is true }
\end{equation}

\section{Related work}
\label{sec:related}
The research in the field of fake news detection has been intense in recent years. However, most of the work in the area is focused on the idea of studying and detecting the hoaxes on its main spreading channel: social media. Examples of the above are ~\cite{FakeNewsDetection1} or~\cite{FakeNewsDetection2}, where the probability of a given post to be false is studied using its own characteristics such as likes, followers, shares, etc., through classical machine learning methods (classification trees, SVM, ...). Applying these kind of approximations, the best results are obtained in~\cite{fakenewscite-093-svm}, where click-bait news are detected achieving results of $93\%$ of accuracy.

Other works like~\cite{FakeNewsDetection3} use graph-based approximations for studying the relations between users who share news and the path which the shared content follows in order to stop it in order to mitigate its potential deception effects.

Although the general trend is to analyze the way hoaxes are spread, other alternatives  focused on the analysis of the content of the news the have begun to appear. In~\cite{FakeNewsDetection4}, besides of the user features who shares the news, the text is used to discriminate fake news. On the other hand, in~\cite{FakeNewsDetection5}, the statements in an article are studied in order to detect false facts in the content.

Regarding the use of modern deep learning algorithms, the company \textit{Fabula.ai}, recently acquired by Twitter, uses a method which takes into account both the content of news and the features extracted from the social networks, achieving results of $0.93$ of AUC~\cite{fabulaai}.

In~\cite{fakenewscite-095-deeplearning} the performance of several algorithms (both classic and deep learning) is compared categorizing news among the categories of true and fake, achieving results of $95\%$ of accuracy.

Finally, using only the content of the news~\cite{TI-CNN}, a convolutional neural network based technique is proposed to detect fake news using the titles and the heading image, obtaining results of $92\%$ of accuracy.
\section{Data}

The dataset used in this work has been built in~\cite{TI-CNN}. It is composed of $20015$ news articles labeled as fake or true which are gathered from two sources. On the one hand, the fake ones come from the dataset Getting Real About Fake News~\cite{gettingreal-kaggle} while the real ones have been obtained from well-known sources such as The New York Times or The Washington Post.

For each article, the dataset contains several features like titles, contents, publisher details, authors or image URLs. However, in this work, only the titles and contents were used. 

The election of this dataset is based on the premise that it has been used previously in other work with good results, so it can be assumed that the data quality of the news is enough to train neural network based models such as the proposed in this work.

Besides TI-CNN dataset, the Fake News Corpus~\cite{fakenews-github} was used for testing and fine tuning the models. This one is composed  by millions of online articles classified in several categories like \textit{bias}, \textit{clickbait}, \textit{fake}, \textit{political} or \textit{true} which were merged accordingly in order to obtain the same labels as the TI-CNN dataset (true and fake).

\subsection{Data transformations}

The process of cleaning the data was trivial as the TI-CNN dataset is well curated. In order to use it for this work, two versions of the dataset were created, as the architecture based on BERT needs a different format (the reason will be explained in the following sections).

The first version consists of the title and body texts of the articles transformed to integer vectors where the position in the vector is the ID of a token in a given dictionary (Word2Vec).

As the input size of a neural network needs to be fixed, a number of words for each title and body needed to be determined. In order to find this maximum number $l$, the number of token distribution   in both titles and contents was studied so as to find a $l$ sufficiently small to be computationally cheap to run the network and sufficiently large to truncate the minimum possible amount of text. Our approximation was to calculate $l$  as $l=\mu + 2\sigma$, obtaining $13$ for the title texts and $1606$ for the bodies. With this numbers, only around $2.5\%$ of the articles was truncated. Also, the vectors which were shorter that the $l$ lenghts, were padded with zeroes.

On the other hand the second version, specific to be used in the BERT based architecture, it's simply composed of the title and body texts concatenated as strings.
\section{Architectures}
This section presents the proposed architectures and gives a brief intuition of the foundations of each one.

Three different neural architectures were created, two of them from scratch and one based on the   BERT language model, created by Google \cite{BERT}. In the next subsections, each one of them is further explained.

\subsection{LSTM based}

This architecture is based on LSTM cells which are a type of recurrent neurons that have probed to give very interesting results in problems related to sequence modeling as they have the capability to ``remember'' information from the past.

The LSTM units are composed of several gates in charge of maintaining a hidden cell state which allows them to mitigate the vanishing gradient problem and, therefore, gives them the ability to remember more distant information in the past than vanilla recurrent units. This feature it's important in the context of NLP since the words from the past often influence the current ones.

More exactly, the architecture uses bidirectional LSTM layers, in which the sequence (ie. the text) is fed forwards and backwards. This decision is based on the intuition that in language, future words modify the meaning of the ones in the present. For example, polysemous words such as \textit{bank}, \textit{mouse} or \textit{book} show that its context is needed in order to model their meaning. 

\begin{figure}[!hbt]
	\centering
	\includegraphics[width=0.9\textwidth]{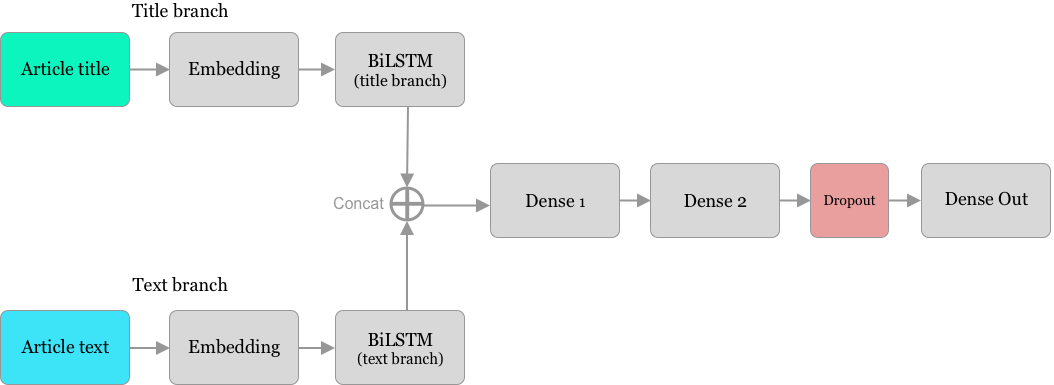}
	\caption{LSTM architecture diagram}
	\label{fig:lstm-arch}
\end{figure}

The diagram of the network is shown in Figure \ref{fig:lstm-arch}. As it can be seen, it is composed of two different branches which merge. One of them is responsible for generating a representation for the title of the article and the other is in charge of the content. Later in the network, these representations are merged and classified in one of the two possible categories (true or fake).

The input (title and content) must be represented as integer vectors, where each entry $e$ is the $i-$th index of a word in a Word2Vec dictionary. The embedding layers were initialized with a pre-trained Word2Vec weights and fixed so that they didn't change during the training.  The goal of this layers is to transform the input vector into a $lxM$ matrix where $l$ is the number of words in the text and $M$ is the dimension of each embedding vector (300 in the case of this work). As it was already mentioned, the $l$ dimension is always the same since in the data curation phase, a padding layer was added to the text in order to obtain a fixed length.  

\subsection{Convolucional based}

Other type of networks known as Convolutional Neural Networks (CNN) have reached the state of the art in almost every computer vision problem, but also, they have demonstrated a great potential in several NLP tasks~\cite{deep-learning-nlp-trends} such as named entity recognition (NER), Part of Speech (POS) Tagging or Natural Language Understanding (NLU).

This kind of networks work by applying a series of filters to their input. These filters are $N-$dimensional matrices which are slided (convoluted) over the input, which means that a series of matrix multiplications are executed in chunks over that input. After training the network, the filters produce activations (known as \textit{feature maps}) where certain patterns are detected (for example, in images, these are borders, figures, patterns, etc.). Additionally another operation which reduces the dimensionality of the output can be applied, called \textit{pooling}, which consist of selecting blocks of $n$ elements from the convolution operation output and performing some operation like the average (Average Pooling) or the maximum (Max Pooling) on each block.

\begin{figure}[!hbt]
	\centering
	\includegraphics[width=0.9\textwidth]{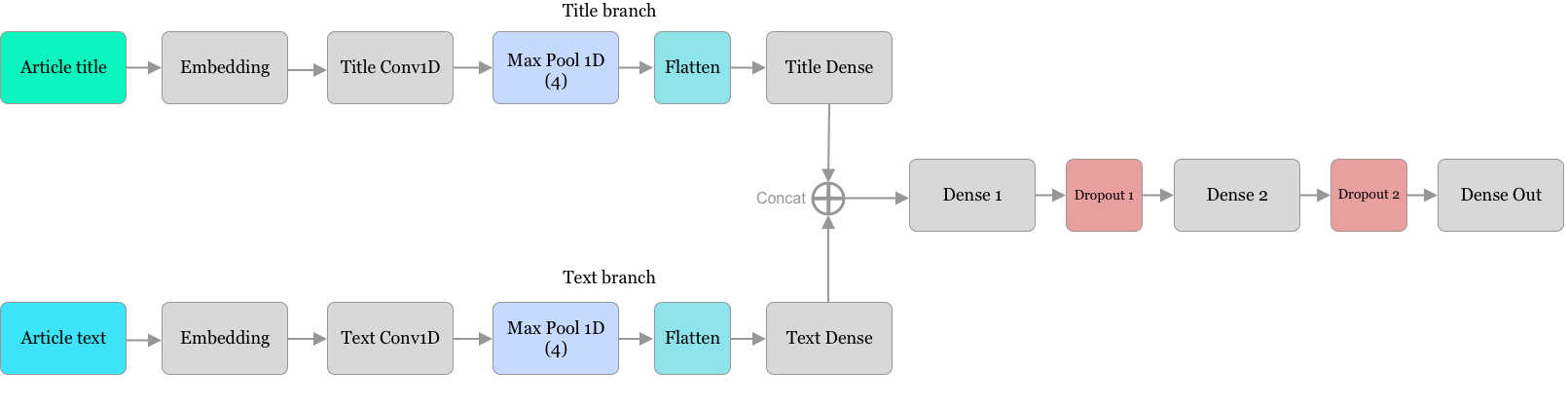}
	\caption{CNN architecture diagram}
	\label{fig:conv-arch}
\end{figure}

Figure~\ref{fig:conv-arch} depicts the diagram of the network. As the LSTM based architecture, this one is also composed of two branches which merge the title and content representations, using this merged features to classify the articles between \textit{true} or \textit{fake} with the same embedding mechanism as the above. The main difference comes from the fact that this architecture uses convolutional layers to analyze different texts. 

\begin{figure}[!hbt]
	\centering
	\includegraphics[width=0.4\textwidth]{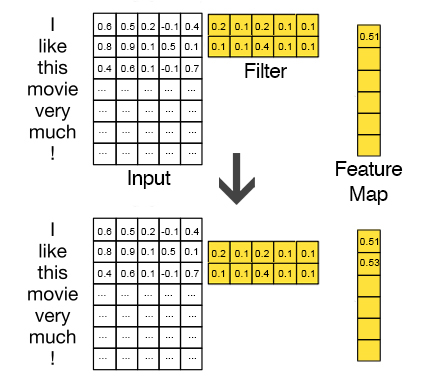}
	\caption{Convolutional operation}
	\label{fig:conv-example}
\end{figure}

The intuition behind this comes from the fact that, after having obtained the word embeddings (see figure~\ref{fig:conv-example}), the content to analyze is similar to an image, i.e a $NxM$ matrix where $N$ is the number of words and $M$ is the embedding dimension. Thus, when a convolutional layer is applied, it is possible to detect the different features defining an article in the same way as border or shapes are in an image. 

However, the main flaw of this type of networks when applied to NLP tasks is that they can't maintain the context from too far in the past, which can make them inferior modeling long term relationships between words compared to a LSTM network.

\subsection{BERT}

In recent years, a huge number of improvements have been made in the field of NLP thanks to deep learning. Most of the recent ones are based on a special type or architecture known as ``transformer''.

\subsubsection{Transformers}
This type of architecture was firstly proposed in~\cite{attention-all-u-need}. It's main goal is, given an input sequence, to transform it into another. The architecture uses ``attention mechanisms'' (also proposed in this paper), which are responsible of determining the most relevant parts of the input sequence. This way, better language representations are created because longer relationships in the sequence can be captured, usually further longer than with LSTM neurons despite of being more  computationally efficient as the operations applied to the input the are simpler.

A transformer is based on the idea of having two pieces: an encoder and a decoder. The encoder creates a representation of a given input in a different dimensional space and then, the decoder takes that representation and generates other sequence. This strategy is called ``encoder-decoder'' and is widely used in tasks like text summarization or machine translation.

\begin{figure}[!hbt]
	\centering
	\includegraphics[width=0.5\textwidth]{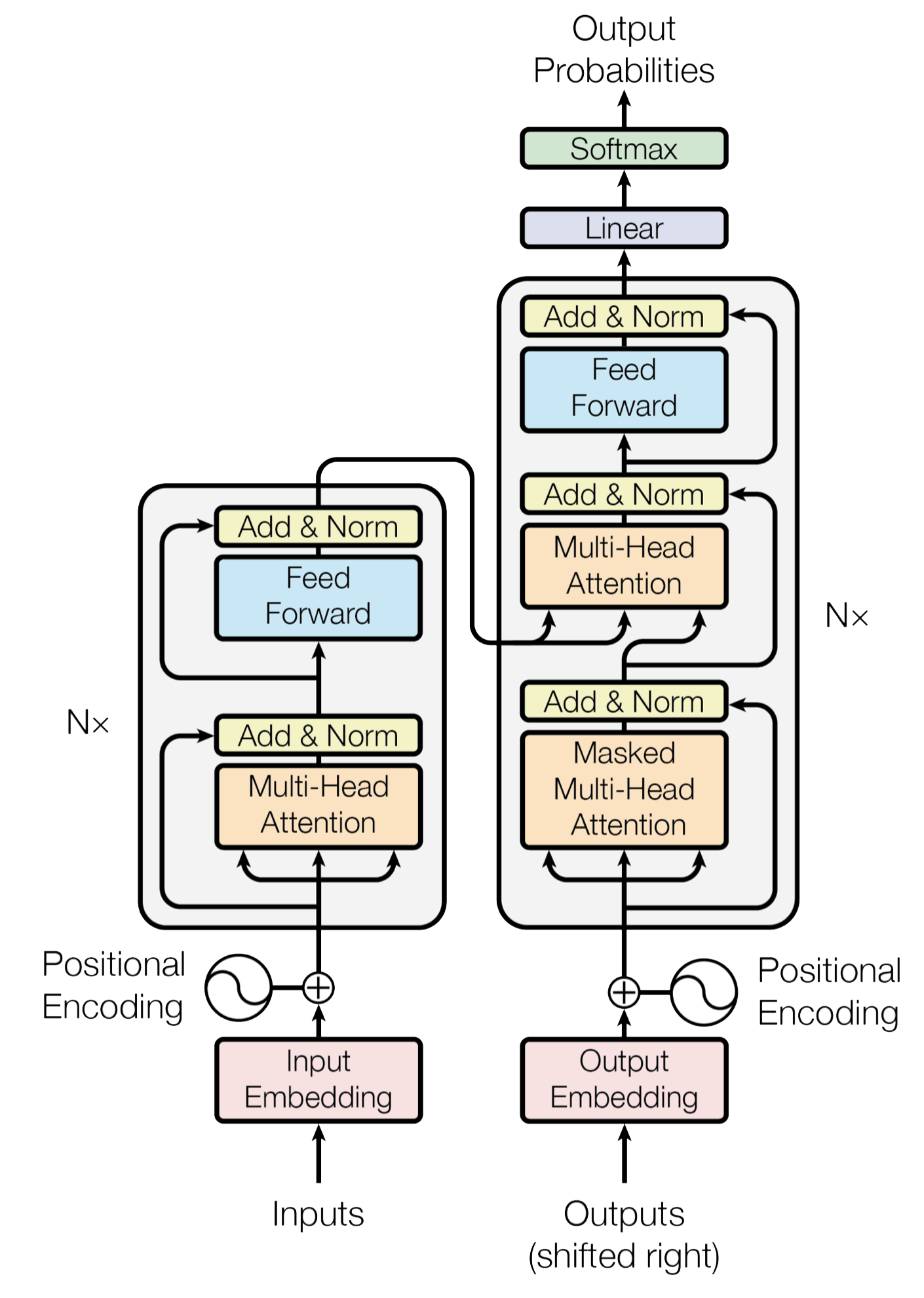}
    \caption{Transformer block diagram. Source:~\cite{attention-all-u-need}}
    \label{fig:transformer}
\end{figure}

A diagram of the transformer architecture is shown in the figure~\ref{fig:transformer}, where the left part corresponds to the encoder block and the right one to the decoder.

Each transformer block uses a ``self-attention'' system which is in charged of choosing the most relevant parts of the input sequence. This system works by operating three matrices: $Q, K$ and $V$ (Query, Key and Value, respectively), which represent an abstraction to calculate the attention matrix, $Z$ (equation~\ref{eq:attention_op}). These three matrices are learnt through in the training phase of the network


After obtaining those matrices the $Z$ matriz can be calculated as shown in equation~\ref{eq:attention_op}, where $d_k$ is the chosen dimension of each key vector (i.e the number of columns in $K$). In the original work, this value corresponds to $d_k=64$.

\begin{equation}
    Z=softmax(\frac{QK^T}{\sqrt{d_k}}V)
	\label{eq:attention_op}
\end{equation}

Also, in the figure~\ref{fig:transformer}, several ``Multi-Head Attention'' blocks can be seen. These simply repeat the attention operation explained above $n$ times, obtaining $n$ attention matrices, which are concatenated and multiplied by other matrix, $W_O$, in order to obtain an output that is fed to the normalization block (``Add \& Norm'', en the figure). .


Besides the Multi-Headed system, the researchers also proposed the use of skip-connections, in such a way an identity signal could be transmited to deeper layers, improving the learning process.


\subsubsection{BERT based}

BERT~\cite{BERT} is a language model created by researchers at Google which is based on transformers. Roughly, it is composed of several stacked transformer encoder blocks.


The strategy to follow with BERT falls under transer learning. BERT is provided already pretrained on a large text corpora  (books, Wikipedia, etc.) with the aim that the final user performs a fine-tunning phase to adapt the model to his specific problem.


Google provides several pretrained models. In their work they present  variants of the architecture:``BASE'' and ``LARGE'' which differ in their size since the first one uses 12 blocks and the second 24 blocks. Due to computational power constraints, in the current work the ``BASE'' version has been used. This is also the approach followed in the original publication. 

\begin{figure}[!htb]
	\centering
	\begin{minipage}[b][][b]{.5\textwidth}
	  \centering
      \includegraphics[width=0.6\textwidth]{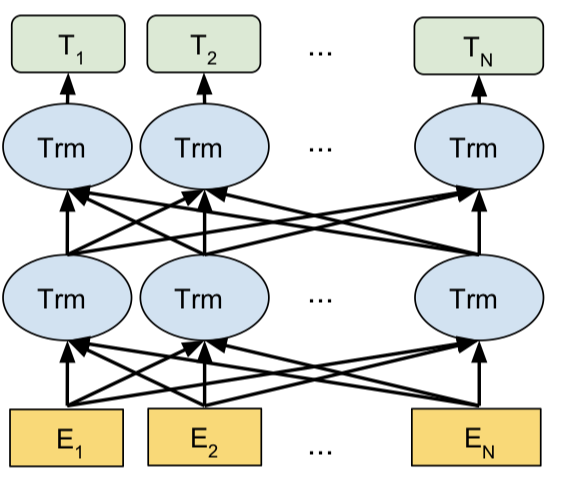}
      \caption{BERT\@. Source~\cite{BERT}}
      \label{fig:bert-arch}
	\end{minipage}%
	\begin{minipage}[b][][b]{.5\textwidth}
	  \centering
      \includegraphics[width=\textwidth]{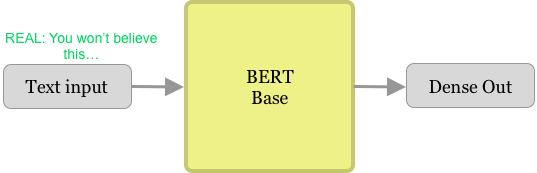}
      \caption{BERT based architecture diagram}
      \label{fig:fine-tuned-bert}
  \end{minipage}
\end{figure}

The adaptation included in this work consists on adding an extra layer to the model provided by Google (see figure~\ref{fig:fine-tuned-bert}). 
This layer is a fully connected layer with a sigmoid as activation function plus a  \textit{softmax} function on top to allow the interpretation of the result as a probability. 

For simplicity in the implementation and the possible future in which the model is required to classify articles in a broader set of categories, the number of neurons in this last layer can be changed as a function of the number of classes in the classification problem. For this binary classification problem (true/false) the number of output neurons is two.

The BERT input data format is different from the ones used for the other two architectures since it is based only on text strings. The word tokenization and separation processes are already included in the input data function for this model. 

The word tokenization follows a strategy called WordPiece. This considers the words as combinations of some more basic tokens joined together. For example, \textit{doing} would be formed by joining \textit{do} and \textit{ing}. By separating the tokens like that, the available lexicon is largely increased, minimizing the potential number of OOV errors (Out Of Vocabulary).

As BERT admits only one input vector, the title and the article body were concatenated before feeding in to the model.

\section{Experiments}

In the first part of this section the hyperparameter tuning method used is described in detail. In the last part the training process of the architectures will be explained.

\subsection{Hyperparameter tuning}

The optimal hyperparameters of the two models created from scratch, shown in tables~\ref{table:hiperpams-lstm} and~\ref{table:hiperpams-cnn}, were found using the Bayesian optimization technique whilst the BERT architecture was fine tuned using the hyperparameters recommended by Google.

This technique aims to estimate the loss space of a model in function of its hyperparameters. The main difference with the most common techniques like \textit{random search} or \textit{grid search} is that the exploration is made only in the most interesting areas of the search space, making it far more efficient.

The classical Bayesian optimization uses a Gaussian Process to estimate the search space and an Acquisition Function (``Expected improvement'', $EI$ in this work, see \ref{ei-def2}) to find the areas which potentially provide better results than the already explored (ie. the points which maximize $EI$).

\begin{equation}
	EI(x) = \cases{
				\mu(x)-f(\hat{x})) \Phi{Z} + \sigma(x) \phi(x) & if $\sigma(x)$ > 0 \cr
				0 & if $\sigma(x)$ = 0 }
	\label{ei-def2}
\end{equation}

The choose of the optimizer was made manually by repeating the Bayesian optimization phase for several of them. Nonetheless, we  found that the best performance in all models was achieved using the SGD optimizer.

On the other hand, due to computational power restrictions, the hyperparameters and optimizer chosen for the BERT architecture were the same as the recommended by Google. Owing to that, we suppose that a more intensive fine tuning on this architecture could lead to further better results.

\begin{table}[!hbt]
\centering
\begin{tabular}{lc}
			\textbf{Hyperparameter} & \textbf{Optimum value} \\
			\hline 
			Learning Rate &  0.264\\ 
			Momentum & 0.082 \\ 
			Dropout & 0.106 \\ 
			Dense 1 Units & 73 \\ 
			Dense 2 Units  & 24 \\ 
			Title Branch LSTM Units & 46 \\ 
			Content Branch LSTM Units & 231 \\ 
		\end{tabular} 
		\caption{Optimal hyperparameters for the LSTM architecture}
		\label{table:hiperpams-lstm}
\end{table}

\begin{table}[!hbt]
	\centering
	\begin{tabular}{lc}
		\textbf{Hyperparameter} & \textbf{Optimum value} \\ 
		\hline 
		Learning Rate &  0.240 \\ 
		Momentum & 0.303 \\ 
		Dropout 1 & 0.110 \\ 
		Dropout 2 & 0.159 \\ 
		Title Conv1D filters & 8 \\ 
		Text Conv1D filters & 42 \\ 
		Title branch Dense units & 6 \\ 
		Text branch Dense units & 34 \\ 
	\end{tabular} 
	\caption{Optimal hyperparameters for the CNN architecture}
	\label{table:hiperpams-cnn}
\end{table}

\subsection{Training conditions}

All models were trained using the hyperparameters explained above with a holdout strategy of $70\%$ of the data for training, $30\%$ for testing and $30\%$ of the training set dedicated to validation during training.

In the architectures built from scratch (Convolutional and LSTM) the optimizer SGD with momentum was used with the optimal parameters obtained during the Bayesian optimization phase. On the other hand, for the architecture based on BERT, Adam was used with the parameters proposed by Google in its original work. 

The loss function chosen for the three models was the ``binary crossentropy'', defined in equation~\ref{eq:binary-xentropy}.

\begin{equation}
    BCE = -y log(p) - (1-y) log(1-p)
    \label{eq:binary-xentropy}
\end{equation}

Besides the regularization measures taken into account (weight initialization and dropout), also \textit{early stopping} was used in order to automatically stop the training when the validation accuracy stops improving.

\section{Results}
This section exposes the results obtained by each one of the three trained models.

\subsection{LSTM model}

\begin{figure}[!hbt]
    \centering
	\includegraphics[width=.36\textwidth]{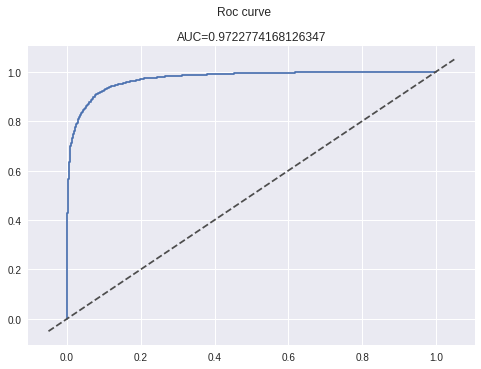}
    \caption{ROC curve of LSTM model over the test dataset.}
    \label{fig:lstm-binary-train}
\end{figure}

The LSTM based model was trained over 5 epochs, obtaining an accuracy of 0.91 on both the validation and the test set. Figure~\ref{fig:lstm-binary-train} shows the ROC curve of the model evaluated on the test partition, obtaining an AUC of 0.97.

\subsection{Convolutional model}
\begin{figure}[!hbt]
    \centering
	\includegraphics[width=0.36\textwidth]{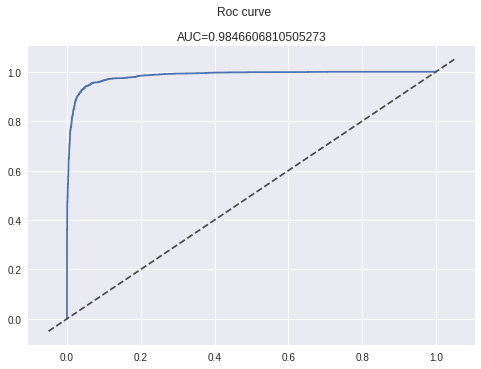}
    \caption{ROC curve of CNN model over the test dataset.}
    \label{fig:conv-binary-train}
\end{figure}

The CNN based model  was trained over 4 epochs, reaching an accuracy of $0.937$ on the validation partition and $0.94$ on the test set. Also, the AUC obatined after evaluating it on the test set was 0.98, as shown in figure~\ref{fig:conv-binary-train}.


\subsection{BERT}
\begin{figure}[!hbt]
	\centering
	\includegraphics[width=0.36\textwidth]{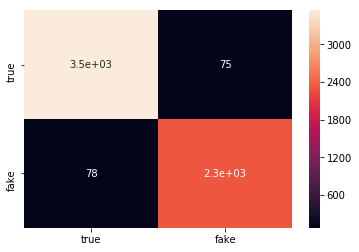}
    \caption{Confusion matrix over the test dataset}
    \label{fig:bert-binary-train}
\end{figure}

This training ran for three epochs with the hyperparameters recommended by Google (\textit{learning rate } = $3x10^{-5}$ and \textit{batch\_size}=32 ). Afterwards, the model was evaluated over the test fold obtaining an accuracy of 0.98 and a F1 metric of 0.97. In addition, the figure~\ref{fig:bert-binary-train} shows the confusion matrix obtained. 

%

\section{Conclusions}

The increasing number of fake news online are a danger for societies and governments as has been already probed. Moreover, the yearly increasing number of hoaxes and the emergence of the artificial intelligence trained to generate texts denote the need of strong systems oriented to discriminate the deceptions.

In this work, we have proposed three novel architectures applied to textual analysis for fake news detection. Two of those created, optimized and trained from scratch and the last one fine tuned from BERT, a pre-trained language model which achieved state of the art results in a great number of NLP tasks.

Although there isn't any benchmark available aimed at evaluating the task of fake news detection, the models here presented outperform the results in the original work which compiled the dataset used~\cite{TI-CNN} ($93\%$ of accuracy) and get superior metrics compared to all the other related ones (see Section \ref{sec:related}). This allow us to presume that it's possible to train neural networks focused on detecting fake news merely using textual features but also that results at the state of the art level can be reached through the strategies proposed.


The experience gained during the development of this models allows us to state that the use of deep learning models for this task can be potentially beneficial for a wide range of actors, from social network companies to the final user in order to mitigate the increasing deceptions on the Internet.

Regarding the future improvement of these models, firstly, it is mandatory to collect more data, specially from a recent period of time. This is also proposed by the researches who compiled the TI-CNN dataset, as the news there are mostly obtained during the US electoral campaign. In order to accomplish the above, a system to automatically collect quality news should be developed. Finally, with the aim that these models can be used by the people, some method of serving them to users is also necessary (integration with social networks, browser extensions, mobile apps...). With these lines of work in mind we expect that our system can be ready to be successfully used in the real world.

\bibliographystyle{unsrt} 
\bibliography{bibliografia}  

\end{document}